%% file: root.tex
\documentclass{article}  % Comment this line out if you need a4paper
\usepackage{iclr2026_conference,times}
\iclrfinalcopy

\usepackage{listings}
\usepackage[utf8]{inputenc} % allow utf-8 input
\usepackage[T1]{fontenc}    % use 8-bit T1 fonts
\usepackage{hyperref}       % hyperlinks
\usepackage{url}            % simple URL typesetting
\usepackage{booktabs}       % professional-quality tables
\usepackage{amsfonts}       % blackboard math symbols
\usepackage{nicefrac}       % compact symbols for 1/2, etc.
\usepackage{microtype}      % microtypography
\usepackage{xcolor}         % colors
\usepackage{comment}
\usepackage{pifont}
\usepackage{wrapfig}

\usepackage{xspace}
\usepackage{makecell}
\usepackage[inkscapearea=page, inkscapepath=figures/generated, inkscapelatex=false]{svg}
\usepackage{caption}
\usepackage{algorithm}
\usepackage{algpseudocode}
\usepackage{graphicx}
\usepackage{multirow}
\usepackage{todonotes}
\usepackage[nolist]{acronym}
\usepackage{amsmath, amssymb}
\acrodef{ai}[AI]{artificial intelligence}
\acrodef{ann}[ANN]{artificial neural network}
\acrodef{llm}[LLM]{Large Language Model}
\acrodef{cad}[CAD]{Computer-Aided Design}
\acrodef{cot}[CoT]{Chain-of-Thought}
\acrodef{gnn}[GNN]{Graph Neural Network}
\acrodef{sam}[SAM]{Segment Anything Model}
\acrodef{api}[API]{Application Programming Interface}
\acrodef{pddl}[PDDL]{Planning Domain Definition Language}
\acrodef{lrm}[LRM]{Large Reasoning Model}
\acrodef{tamp}[TAMP]{Task and Motion Planning}
\acrodef{ipc}[IPC]{International Planning Competition}
\acrodef{ew}[EW]{Exploration Walk}
\acrodef{cands}[C\&S]{cluster-and-search}

\newif\ifnotanonym
\notanonymfalse
\ifnotanonym
\newcommand{\linktowebsite}{\url{https://claudius-kienle.github.io/lodge}}
\fi

\title{
Learning Hierarchical Domain Models through Environment-Grounded Interaction
}

\author{%
  Claudius Kienle \\
  Intelligent Autonomous Systems Group\\
  TU Darmstadt\\
  Germany \\
  \texttt{claudius.kienle@tu-darmstadt.de} \\
  \And
  Benjamin Alt \\
  AICOR Insitute for Artificial Intelligence \\
  University of Bremen \\
  Germany \\
  \texttt{benjamin.alt@uni-bremen.de} \\
  \AND
  Oleg Arenz \\
  Intelligent Autonomous Systems Group\\
  TU Darmstadt\\
  Germany \\
  \texttt{oleg.arenz@tu-darmstadt.de} \\
  \And
  Jan Peters \\
  Intelligent Autonomous Systems Group\\
  TU Darmstadt\\
  Germany \\
  \texttt{jan.peters@tu-darmstadt.de} \\
}

\begin{document}

\maketitle

\begin{abstract}
    Domain models enable autonomous agents to solve long-horizon tasks by producing interpretable plans. However, in open-world environments, a single general domain model cannot capture the variety of tasks, so agents must generate suitable task-specific models on the fly.
    \acp{llm}, with their implicit common knowledge, can generate such domains, but suffer from high error rates that limit their applicability. Hence, related work relies on extensive human feedback or prior knowledge, which undermines autonomous, open-world deployment. In this work, we propose LODGE, a framework for autonomous domain learning from \acp{llm} and environment grounding. LODGE builds on hierarchical abstractions and automated simulations to identify and correct inconsistencies between abstraction layers and between the model and environment. Our framework is task-agnostic, as it generates predicates, operators, and their preconditions and effects, while only assuming access to a simulator and a set of generic, executable low-level skills. 
    %Autonomous agents in open-world environments must  solve long-horizon tasks such as table setting or warehouse operations. The resulting plans are structured, interpretable, and grounded in explicit domain models. However, in open-world environments we can not resort to a single ground-truth domain model, but have to learn and adapt a context-specific domain. \acp{llm}, with their implicit common knowledge, offer the potential to autonomously propose predicates, operators, and their preconditions and effects for context-specific domains. However, as the outputs of the LLM often contain errors, requiring extensive human feedback or prior knowledge.
    %We propose LODGE, an autonomous method that learns hierarchical domain models by combining \acp{llm} and environment grounding.
    %To address the complexity of learning and planning for large domains, we introduce hierarchical decomposition where abstract operators are progressively refined into subdomains. 
    %We learn predicate classifiers for newly invented predicates to detect misalignments between the symbolic state of a plan and its execution in the environment, and introduce a global recovery module that adapts the domain model to correct occurred misalignments.
    Experiments on two \ac{ipc} domains and a robotic assembly domain show that LODGE yields more accurate domain models and higher task success than existing methods, requiring remarkably few environment interactions and no human feedback or demonstrations.
    \ifnotanonym
        Resources, videos and detailed experiment results are available at \linktowebsite.
    \fi
\end{abstract}

\section{Introduction}
\label{sec:introduction}
A central goal in robotics is to enable autonomous agents to operate in open-world environments and solve long-horizon tasks such as setting a table \citep{sun_semantic_2019} or retrieving and stowing packages across a warehouse \citep{bernardo_survey_2022}. Planning is essential in this context, as it produces structured and interpretable plans grounded in explicit domain models, in contrast to opaque VLAs \citep{firoozi_foundation_2025}.
The domain model describes the world through objects, predicates defining affordances and relations, and operators specifying available actions.
However, open-world environments make it impossible to assume a single ground-truth domain model: The relevant objects, relations, and actions differ across tasks and settings. Even if constructing a single, universal domain were feasible, planning within it would be computationally intractable. Hence, learning context-specific domain models on the fly is crucial for explicit, interpretable open-world reasoning. Significant work has addressed learning domain models from demonstrations \citep{silver_learning_2021}. 
However, to be practical for open-world settings, domain models must be learned from limited data, since collecting extensive demonstrations or annotations for every new setting is infeasible.
Instead, \acp{llm} can facilitate autonomous online-learning of context-specific domain modes by retrieving relevant knowledge.
However, their outputs are not reliable: forgotten or hallucinated preconditions and effects
% , or incorrectly grounded predicates
may result in domains that are unsolvable, not executable by the robot, or misaligned with the intended goal.
Hence, prior works rely on verbose and task-specific skill annotations~\citep{mahdavi_leveraging_2024} or significant human feedback~\citep{guan_leveraging_2023} to improve the correctness of LLM-generated domains, but thereby also limit their applicability for online, open-world domain learning.

%Previous work use world knowledge from \acp{llm} to construct domain model from relatively little data. However, current \ac{llm}-based approaches still face practical limitations: Partially automated systems require significant human feedback \citep{guan_leveraging_2023}, while fully automated methods depend on verbose skill annotations, partially known domain models \cite{mahdavi_leveraging_2024}, which prevents their applicability to open-world problems.

To address these challenges, we propose LODGE, a fully autonomous and task-generic method for learning domain models. 
LODGE learns domain models without trusted knowledge sources by enforcing consistency across layers of a hierarchical domain model, between the domain model and a robot simulation, and across interacting LLM and VLM-based modules.
LODGE leverages an \ac{llm} to generate an initial domain model, which it iteratively refines and adapts in a targeted manner while solving tasks using the domain model in the environment. We show that our method requires remarkably few environment interactions to construct \ac{pddl} domain models that are more accurate than related work.
The difficulty of learning domain models with \acp{llm} increases with the number of operators and predicates \citep{kambhampati_llms_2024}.
To address this, we propose to learn hierarchical domain models. LODGE first generates an abstract domain model and plan that it subsequently decomposes into subdomains and plans.
Hierarchical decomposition enables partial re-planning of subplans while maintaining alignment across hierarchy levels. Unlike approaches that propose sub-goals \citep{liu_delta_2025}, hierarchical domain models do not suffer from detrimental sub-goal interactions \citep{kambhampati_llms_2024}.
To avoid relying on predefined task-specific predicates, we propose an algorithm for inventing new predicates and corresponding classifiers to ground the symbolic state of the environment from the low-level continuous state.
We show how the accuracy of learned classifiers can be improved by autonomously optimizing the classifiers' hyperparameters (e.g. tolerances) and by \ac{llm} self-refinement based on a dataset of past environment interactions, achieving higher accuracy than related methods.
Lastly, we propose a \textit{global recovery module} that detects misalignments between the symbolic state of the environment and the planner's symbolic state. Upon misalignment, it analyzes the past domain learning and trigger a targeted recovery step to restore alignment with the observed symbolic state and recover domain learning.
With that, our contributions are as follows:

\textbf{Domain learning from planning feedback:}
By combining \acp{llm} with predicate classifiers, we detect misalignments between the symbolic planner's state and the observed environment state. With the \textit{global recovery module}, we show how domain operators can be refined in a targeted manner. This proposes the first automated method to learn domain models from few environment interactions without human feedback, enabling use in open-world environments.

\textbf{Hierarchical domain models:}
We introduce a hierarchical approach to domain learning in which abstract operators are progressively decomposed into reusable subdomains. This enables the distribution of domain model complexity across multiple abstraction levels, which simplifies learning and reduces computational planning complexity.

\textbf{Predicate invention and online classifier learning:}
We introduce a method to invent new predicates and learn their classifiers from pseudo-labeled environment interactions. This shows that the symbolic state can be extended autonomously, which enables representing and grounding previously unknown aspects of the environment without human feedback.

\section{Related Work}
\label{sec:related_work}

% We combine domain learning with \acp{llm} with environment grounding via predicate classifiers. 
% Below, we first overview planning with \acp{llm}, then review prior work on \acp{llm}-based domain models learning, and finally discuss research on inventing predicates and learning classifiers.

\paragraph{Planning with \protect\acp{llm}}
Early work used \acp{llm} to generate plans directly from natural language \citep{ichter_as_2023}. \citet{liang_code_2023} proposed hierarchical plan decomposition, enabling \acp{llm} to define and later implement unknown functions. Others explored specifying tasks with formal planning languages \citep{valmeekam_planning_2023} to reduce natural language ambiguity \citep{pallagani2024prospects}. While natural language instructions offer flexibility, \ac{llm} planners lack correctness guarantees, making plan validation difficult. State-of-the-art models still struggle with simple planning due to limited explicit reasoning \citep{valmeekam_planning_2023}.
To address these limitations, hybrid methods combine \acp{llm} with classical planners. One approach uses \acp{llm} to translate natural language instructions into formal goal specifications interpretable by classical planners \citep{liu_llmp_2023}.
The syntax and semantics of problem definitions can be verified externally \citep{howey_val_2004} or via \ac{llm} self-critiques \citep{zhou_isr-llm_2024}.
Unlike classical planners, \ac{llm} planners provide feedback when plans fail \citep{valmeekam_planning_2023}, and plan correctness can be validated externally \citep{zhou_isr-llm_2024}
However, these methods assume a predefined domain model, which is difficult to construct.

\paragraph{Learning Domain Models with \acp{llm}}
Constructing domain models is complex, but crucial for autonomous planning systems. The construction with \acp{llm} is, due to the resulting sample efficiency, particularly promising for open-world domains, but remains underexplored \citep{tantakoun2025llms}. Recent work verifies \ac{llm}-generated domains by self-critique or classical tools \citep{smirnov_generating_2024,ishay_llm_2025}. Complementary work generates multiple domain candidates and evaluates them using similarity metrics or environment interaction \citep{yu2025generating,hu_text2world_2025}.
\citet{guan_leveraging_2023} propose an \ac{llm}-based framework to construct PDDL domain models by first translating predefined skills into action definitions, followed by iterative refinement via human feedback.
\citet{oswald_large_2024} focus on reconstructing high-quality PDDL domains from natural language descriptions. Recent work by \citet{mahdavi_leveraging_2024} avoid human feedback by sampling multiple candidate domains and iteratively refining the one closest to a partially known domain model.

In contrast to these approaches, LODGE learns hierarchical domain models with \acp{llm} and environment grounding, without skill annotations or human feedback.
Additionally, LODGE does not rely on partially known domain models and introduces a targeted refinement of models through environment interaction.

\paragraph{Predicate Invention and Learning Classifiers}
Previous work extracts predicates from demonstration data \citep{keller_neuro-symbolic_2025} or learns predicate classifiers that map the continuous environment state to the planner's symbolic state \citep{migimatsu_grounding_2022}.
Furthermore, predicates can be infered and grounded by constructing scene graphs~\citep{ekpo_verigraph_2024,herzog_domain-conditioned_2025}.
Large pretrained models can also ground predicates from images of the scene \citep{liang_visualpredicator_2025,athalye_pixels_2025}, or from perception data, such as 6D poses \citep{shirai_vision-language_2024}.
Alternatively, \acp{llm} can generate interpretable predicate classifiers in code, e.g., Python, avoiding the need for \acp{llm} during inference \citep{han_interpret_2024}.

Our algorithm for learning predicate classifiers combines recent work of \citet{han_interpret_2024} and \citet{athalye_pixels_2025}. We iteratively refine the constructed predicate classifiers based on past environment interactions labeled with pseudo-labels instead of requiring human feedback, and propose two new techniques to refine predicate classifiers based on a pseudo-labeled dataset.

\section{Notation and Problem Statement}

We consider an environment $f$ (real or simulated) with a continuous state $x$, where each object $o \in \mathcal{O}$ is mapped to real-valued vectors, e.g. its 6D pose. An actor can execute a skill $\pi$ in the environment, which is parameterized by a set of variables $\mathcal{V}$ that can be grounded with objects $o_i \in \mathcal{O}$ existing in the environment. We denote a sequence $o_1, \dots, o_n$ by $o_{1:n}$.
Executing $\pi$ on $f$ transitions the environment into the next state $x_{t+1} = f(x_t, \pi(o_{1:n_\pi}))$.
The domain model $\mathcal{D} = \langle \Psi, \Omega \rangle$ consists of predicates $\Psi$ and operators $\Omega$.
Every predicate $\psi \in \Psi$ encodes a discrete property of the state $x$, such as \textit{grasps(?obj)} or \textit{stacked(?obj1, ?obj2)}. We distinguish between \textit{state-based} predicates that operate exclusively over the object-centric state, and \textit{state-independent} predicates.
Each \textit{state-based} predicate $\psi$ has a corresponding classifier $c_\psi: \mathcal{O} \times \mathcal{X} \to \{\text{true}, \text{false}\}$. 
$c_{\underline\psi}(x)=c_\psi(o_{1:n_\psi}, x)$ evaluates whether predicate $\psi$ with objects $o_{1:n_\psi}$ (ground atom $\underline\psi$) holds in low-level state $x$. \textit{State-independent} predicates encode information that cannot be retrieved from the low-level state $x$, like object affordances (e.g. \textit{graspable(?obj)}).
We denote an evaluation of all predicates $\Psi$ on state $x$ as symbolic state $s = \{\underline\psi: c_{\underline\psi}(x) = \text{true}, \forall \psi \in \Psi\}$.
Every operator $\omega \in \Omega$ consists of preconditions $\text{pre}(\mathcal{\omega})$ that define when $\omega$ can be executed, and effects $\text{eff}(\omega) = \langle EFF^+, EFF^- \rangle$ that define the add and delete effects of $\omega$ on the symbolic state $s$.
We refer to an operator instantiated with objects $o_{1:n_\omega}$ for each variable $\mathcal{V}$ as action $a=\omega(o_{1:n_\omega})$.
Additionally, $\text{eff}(a_{1:n})$ defines the state change resulting from applying plan $a_{1:n}$.
We distinguish between skills $\pi$ and actions $a$: Actions are part of a high-level task plan, while skills are the low-level primitives the actor can execute on the continuous state $x$. An action $a$ is mapped to one or more skills $\pi$. Hierarchy levels are denoted with a superscript, where $D^{(0)}$ is the domain model at the upmost hierarchy level.
A goal $g = \langle g^+, g^- \rangle$ is achieved if, for every $\underline{\psi} \in g^+$, $c_{\underline\psi}(x)$ is true, and for every $\underline{\psi} \in g^-$, $c_{\underline\psi}(x)$ is false (see \citet{silver_predicate_2023}).

Our objective is to learn domain models for long-horizon sequential tasks that require geometric and symbolic reasoning.
We assume a deterministic environment $f$ and a list of all objects $\mathcal{O}$, which can be retrieved, for example, using segmentation~\citep{kirillov_segment_2023}. The environment state $x$ is fully observable.
We assume a library $\Pi$ of skills with their names and parameter signatures.
Executing a skill in the environment either transitions the environment into the next state or returns an exception that occurred during skill execution (e.g. for an invalid parameterization).
Given a task specified by a natural language instruction $I$, we learn a hierarchical domain model that solves that and similar future tasks. Specifically, we construct operators, predicates and predicate classifiers.

\section{METHODS}

We propose LODGE, a method to learn explicit hierarchical \ac{pddl} domain models leveraging \acp{llm} and environment grounding (see Fig.~\ref{fig:architecture}). We propose to \textbf{decompose} (see Sec. \ref{sec:action-decomposition}) domains hierarchically, which distributes the complexity of learning across hierarchy levels and enables faster planning for long-horizon tasks. We ensure the bidirectional \textbf{alignment} (see Sec. \ref{sec:knowledge-preservation}) between hierarchy levels, which is required to unify the subplans at different hierarchy levels to one valid hierarchical plan, avoiding infeasible plans due to sub-goal interactions. Plans are validated against the domain model, and executed in an environment. %to verify the correctness of the domain model against the environment. 
We \textbf{invent predicates and learn classifiers} (see Sec. \ref{sec:learning-predicates}) to estimate the symbolic state of the environment and compare it with the planner's state.
A \textbf{global recovery module} (see Sec.~\ref{sec:motion-verification}) analyzes detected misalignments and suggests how to adapt the model to recover learning.
% between the planner's state and observed environment state, their origin in the past domain learning process and how to recover model learning.

\subsection{Learning Hierarchical Domain Models by Planning}
\label{sec:action-decomposition}

A central challenge in learning symbolic domain models is that initial models are incomplete and contain errors. We address this by incrementally constructing a hierarchical domain model, alternating between domain generation, planning, and environment grounding (see Fig. \ref{fig:architecture} and Alg. \ref{alg:decomposition}).

Given a natural language instruction $I$, skill library $\Pi$, a start state $x_0$, and an optional set of initial predicates $\Psi_{\text{init}}$, LODGE first induces a preliminary domain model $\mathcal{D}^{(0)}$ and goal state $g^{(0)}$:
\[
    \mathcal{D}^{(0)}, g^{(0)} = \text{LLM}_{\text{domain}}(I, c_\Psi(x_0), \Psi_{\text{init}}, \Pi).
\]
The induced domain model $\mathcal{D}^{(0)}$ introduced new operators $\omega_i$ and predicates $\psi_j$, which are automatically classified as either \textit{state-based} or \textit{state-independent}. Only \textit{state-based} predicates are considered for classifiers learning (see Sec. \ref{sec:learning-predicates}) and Motion Verification (see Sec. \ref{sec:motion-verification}).
The domain $\mathcal{D}^{(0)}$ and goal state $g^{(0)}$ are checked for syntax errors, and corrected with self-refinement \citep{howey_val_2004}.

\paragraph{Abstraction through hierarchical operators}
Instead of generating one operator for each predefined skill $\pi$, we explicitly instruct the \ac{llm} to define high-level operators. We later iteratively decompose these operators into lower-level domain models, forming a hierarchical domain model.
This hierarchical approach provides two key benefits:
(1) Multiple abstraction levels significantly reduce the domain model complexity by reducing the number of operators at one level and their individual complexity, which also improve \ac{llm} accuracy for suggesting plans and reduces the exponential search time of classical planners.
(2) Operators from a higher-level domain can operate on a coarse-grained symbolic state, while decomposition progressively refines state representation. Defining fine-grained predicates, e.g. the gripper position relative to an object, at the top level can be ambiguous and tracking their state complicates the operators (cf. Sec. \ref{subsec:furniture-bench}). 

\paragraph{Plan generation under partial domain models}
LODGE uses domain $\mathcal{D}^{(0)}$ and problem $\mathcal{P}^{(0)} = \langle \mathcal{O}, s_0, g^{(0)} \rangle$ to generate a candidate plan
$a_{1:n}^{(0)} = \text{GenPlan}_{\text{LLM}}(\mathcal{D}^{(0)}, \mathcal{P}^{(0)})$.
At early stages, the classical planner often fails due to an incorrect domain. In such cases, LODGE reverts to an \ac{llm}-generated plan. While this plan may be invalid, it is required to find the misalignment between the domain model and environment during \textit{Motion Verification} (see Sec. \ref{sec:motion-verification}).
In contrast to related work \citep{wong_learning_2023}, we do not validate the plan correctness against the domain model, e.g. with VAL \citep{howey_val_2004}. VAL only detects unsatisfied preconditions of an action, but these can also be caused by missing effects of previous actions. Instead, the plan is verified during \textit{Motion Verification}.

\begin{figure}
    \includesvg[width=\linewidth]{images/pddl-llm-overview.svg}
    \caption{LODGE: Learning hierarchical domain models with environment grounding.}
    \label{fig:architecture}
\end{figure}

\paragraph{Mapping to skills}
Given a plan candidate $a_{1:n}^{(0)}$, we iterate over every action $a^{(0)}_i = \omega(o_{1:n})$
to determine whether the operator $\omega$ must be decomposed or whether it can be realized with a single skill $\pi \in \Pi$ (Alg. \ref{alg:decomposition}).
A translation model maps $\omega$ to a sequence of skills $\pi_{0:T-1} = \text{LLM}_\text{translate}(\omega, \Pi, \mathcal{O})$ parameterized by the operator's arguments.
% \[
%     \pi_{0:T-1} = \text{LLM}_\text{translate}(\omega, \Pi, \mathcal{O}).
% \]
If this \textit{lifted} skill sequence consists of a single skill, we mark $\omega$ as leaf node and start motion verification. We otherwise \textit{decompose} the operator $\omega$ (Fig. \ref{fig:architecture}).

\paragraph{Decomposition}
The goal of decomposing an operator $\omega^{(0)}$ is to construct a lower-level domain $\mathcal{D}^{(1)}$. For example, the operator \textit{grasp(bulb)} can be decomposed into \textit{approach(bulb)}, \textit{close\_gripper(bulb)}, and \textit{lift(bulb)}.
The action $a^{(0)}_i=\omega^{(0)}(o_{1:n_{\omega}})$ in a sequence $a^{(0)}_{1:m}$ causes a state transition from $s_i$ to $s_{i+1}$, where $\text{pre}(\omega^{(0)})$ is a subset of $s_i$ and $\text{eff}(\omega^{(0)})$ describes the state change from $s_i$ to $s_{i+1}$ \citep{howey_val_2004}.
To decompose $\omega^{(0)}$, we define a subproblem $P^{(1)}$ and subdomain $D^{(1)}$ with:
\begin{align}
    \mathcal{P}^{(1)}      &= \langle \mathcal{O}, s_i,  \text{eff}(s_i,s_{i+1}) \rangle, \quad \text{where} \quad
    \text{eff}(s_i, s_{i+1}) = \langle s_{i+1} \setminus s_i, s_i \setminus s_{i+1} \rangle,
    \label{eq:effect-formula} \\
    \mathcal{D}^{(1)} &= \ac{llm}_{\text{decomp}}(\mathcal{P}^{(1)}, \Psi^{(1)}_\text{init}, \Pi, \omega^{(0)}, \pi_{0:T-1}), \nonumber
\end{align}
where $\Psi^{(1)}_\text{init} := \Psi^{(0)}$ provides the initial predicate set.
In contrast to $\text{LLM}_\text{domain}$, the decomposition model $\ac{llm}_\text{decomp}$ conditions on the high-level operator $\omega^{(0)}$ and problem $\mathcal{P}^{(1)}$, rather than on the natural language instruction $I$.
We additionally provide the proposed skill sequence $\pi_{0:T-1}$ generated by $\text{LLM}_\text{translate}$ to align the decomposition with it.
% Importantly, this formulation ensures that decompositions generalize across all instantiations of an operator.
Importantly, once a decomposition $\mathcal{D}^{(1)}$ for $\omega^{(0)}$ is learned, it can be reused for any subsequent action $a_j=\omega^{(0)}(o'_{1:n_\omega})$ without additional \ac{llm} calls. 
% The corresponding subproblem $\mathcal{P}^{(1)}$ can be automatically constructed from the states $s_j$ and $s_{j+1}$.

\subsection{Aligning Levels of Hierarchical Domain Models}
\label{sec:knowledge-preservation}
Hierarchically decomposing the domain model presents two key challenges: (1) preserving knowledge across levels and (2) ensuring alignment between levels to ultimately produce a coherent hierarchical domain model and allow for hierarchical planning.

To preserve knowledge, we retain predicates $\Psi$ and objects $\mathcal{O}$ from one hierarchy level to all lower ones. Retaining the predicates for lower levels is crucial
for aligning them with the environment during motion verification.
%as the motion verification verifies their alignment with the environment state.
However, lower-level predicates are not available at upper levels to encapsulate state tracking.
The decomposition of a high-level operator $\omega$ with predicates $\Psi^{(0)}$, and action $a=\omega(o_{1:n_\omega})$ with effects $\text{eff}(a)$ generates low-level operators, predicates $\Psi^{(1)} \supseteq \Psi^{(0)}$ and a subplan $a_{1:k}$ with joint effects $\text{eff}(a_{1:k})$. To keep $\omega$ aligned with the subplan, the effects $\text{eff}(a)$ must match the effects $\text{eff}(a_{1:k})$ on the upper-level predicates $\Psi^{(0)}$. It holds that $\text{eff}(a) \subseteq \text{eff}(a_{1:k})$, as the goal state of the lower-level problem follows from the action's effects $\text{eff}(a)$.
However, $\text{eff}(a_{1:k})$ can be a superset of $\text{eff}(a)$ such that more predicates in $\Psi^{(0)}$ change than initially assumed.
For example, the high-level action \textit{pick-up(apple)} has effect \textit{grasps(\textit{apple})}. 
Decomposition leads to a joint effect $[$\textit{grasps(\textit{apple})}, \textit{door-open(\textit{drawer})}, \textit{closed-gripper()}$]$, $\text{eff}(a_{1:k})$, introducing two predicates not in $\text{eff}(a)$.
We call these predicates $\text{eff}(a_{1:k}) \setminus \text{eff}(a)$ \textit{overshoots} and \textit{side effects}: \textit{Overshoots} operate solely on objects used in the action ($\forall o \in o_{1:n_\omega}$), while \textit{side effects} operate on at least one object not used in the action; $\exists o \notin o_{1:n_\omega}$.
In this example, \textit{closed-gripper()} is an \textit{overshoot} and \textit{door-open(\textit{drawer})} a \textit{side effect} of the decomposition.

These \textit{misalignments} between hierarchy levels can break plan correctness at the upper level due to sub-goal interactions. After action $a$, the upper-level state reflects $\text{eff}(a)$, but executing the subplan $a_{1:k}$ changes the state by $\text{eff}(a_{1:k})$. Considering the example from above. If action \textit{close-gripper()} follows next with precondition $\textit{not(closed-gripper)}$, the plan appears valid at the upper level but fails after decomposition because the subplan's effects defines $\textit{closed-gripper}$ as true.

We detect misalignments by comparing $\text{eff}(a)$ with $\text{eff}(a_{1:k})$ on the upper-level predicates $\Psi^{(0)}$ after executing the subplan. If misalignments occur, $\ac{llm}_\text{decomp}$ realigns the operator $\omega$ of $a$ with its subplan $a_{1:k}$.
Fixing \textit{overshoots} is simpler than \textit{side effects}, since \textit{side effects} affects objects not used in $\omega$, which requires adding new variables to $\omega$, i.e. \textit{drawer} in the example above. Smaller models like GPT 4o mini handle \textit{overshoots} well but struggle with \textit{side effects}.

\subsection{Inventing Predicates and Learning Classifiers}
\label{sec:learning-predicates}

The definition of predicates $\Psi$ can be
% , depending on the environment, 
non-trivial and can influence learning accurate domain models.
Additionally, robotic environments do not expose symbolic states, requiring a classifier $c_\psi$ to ground predicates in the continuous environment state $x$.
Apart from the initial predicates $\Psi_\text{init}$, LODGE invents predicates during domain learning as needed. We automatically learn predicate classifiers for newly invented \textit{state-based} predicates when grounding is required.

We generate Python-based classifiers with an \ac{llm} as in \citet{han_interpret_2024}, but prompt it to add hyperparameters $\boldsymbol{\theta}$ with plausible default values  $\boldsymbol{\theta}^\text{default}$ as arguments to the function, e.g. `def holding(obj: Part, \textit{xy\_tolerance: float = 0.05})'. The Python code has access to the perceived object poses and the robot end-effector state (see App. \ref{app:py-predicate-grounding}).
Compared to \acp{llm} or VLMs, Python-based classifiers are explicit, interpretable, and significantly more efficient. Additionally, relying on the perception of the environment instead of \acp{llm} avoids hallucinations.
However, the generated code can contain mistakes, i.e. syntax or logical errors.
% Additionally, most classifiers depend on hyperparameters, e.g. a predicate \textit{obj-on-obj} has thresholds that define when objects are overlapping or the maximal gap allowed between objects to be considered `on' (see Fig. 2, \citet{han_interpret_2024}). 
Additionally, most classifiers depend on hyperparameters (see above or Fig. 2, \citet{han_interpret_2024}). 
Robust values for these hyperparameters vary with the system and environment, e.g. the perception noise.
To address these challenges, we propose (1) a method to collect pseudo-labeled interactions and (2) algorithms to optimize the Python-based classifiers on that data.

\paragraph{Data Collection} 
We record all environment transitions $t_i = (x_i, \pi_i, x'_i) \in \mathcal{T}$ during Motion Verification (see Sec. \ref{sec:motion-verification}), consisting of state $x_i$, skill $\pi_i$, and next state $x'_i$.
For all next states $x'_i$ in the transitions, we predict pseudo-labels $s'_i$ similar to \citet{athalye_pixels_2025} (see App. \ref{app:vlm-grounding}). 
% They obtain $s'_i$ from a VLM given the transition $t_i$ and an image of the scene at state $x'_i$. The start state $x_0$ is labeled directly from $x_0$ and the image.
This dataset $\mathcal{D} = \{(x_i, \pi_i, s_i)\}_{i=1}^n$ is used to evaluate and refine the predicate classifiers.
Given a constructed Python classifier $c_\psi$ and dataset $\mathcal{D}$, we evaluate $c_\psi$ by computing the F1 score $\mathcal{F}_{\underline\psi}$ on $\mathcal{D}$ for every ground atom $\underline\psi$. % , i.e. predicate $\psi$ instantiated with objects $o_{1:n}$.
We refine the predicate classifier based on the lowest F1 score of any ground atom.
On a f1 score below $\tau_\text{hp}$, we refine the classifier's hyperparameters $\theta$ for a better evaluation metric via \textit{hyperparameter optimization}.
When the f1 score is below $\tau_\text{llm}$, we improve the code of the classifier via \textit{\ac{llm} self-refinement}. $\tau_\text{llm}$ is generally lower than $\tau_\text{hp}$, suggesting an error in the code of the classifier. We empirically used $\tau_\text{hp} = 0.9$ and $\tau_\text{llm} = 0.6$ for all our experiments.

\paragraph{Hyperparameter Optimization}
We optimize the LLM-given default hyperparameters $\boldsymbol{\theta}^\text{default}$ via gradient-free random search evaluated on the average F1 score
$\mathcal{F}(\boldsymbol\theta) = \langle \mathcal{F}_{\underline\psi}(\boldsymbol{\theta}) \rangle_{\underline\psi}.$
Among all samples that achieve the best F1 score $\mathcal{F}$, 
we first identify the most robust candidates and then select the one closest to $\boldsymbol\theta^\text{default}$.
We evaluate the robustness $R(\boldsymbol\theta)$
of $\boldsymbol\theta$ as the minimal relative change in any hyperparameter $k$  that changes the F1 score $\mathcal{F}(\boldsymbol\theta)$, 
\[
    R(\boldsymbol\theta) = \min_{\boldsymbol\theta' : f(\boldsymbol\theta') \neq f(\boldsymbol\theta)}
    \; \min_{k} \frac{| \theta'_k - \theta_k |}{|\theta^{\text{default}}_k|}.
\]

\paragraph{\ac{llm} Self-Refinement}
We invoke the \ac{llm} to refine the Python classifier. For this, we sample up to three samples from $\mathcal{D}$ on which the Python classifier yields a different classification than the pseudo-label and prompt the \ac{llm} with the current Python classifier and the samples to either refine the code, or leave it as is if it believes the VLM labels are incorrect. Every sample in the prompt is stated by the 6D poses of robot and objects, the classification results of the Python classifier and VLM, and the classification result of the Python classifiers used in the Python code (see App. \ref{app:vlm-grounding}).

\subsection{Verifying Plans, Recovering from Model Errors, and Adapting Learning}
\label{sec:motion-verification}
% $\text{LLM}_\text{domain}$ and $\text{LLM}_\text{decomp}$ generate the domain model based on their world knowledge and provided input. As a consequence, the domain model is highly contingent on the world knowledge of the \ac{llm}. Consequently, modeling errors and the lack of planning capabilities of the \ac{llm} can result in invalid plans. 
To evaluate domain model correctness, we execute every planned skill in simulation and compare the planner's symbolic state change with the observed state change in simulation. A misalignment indicates an error in the learned domain model that we address with a \textit{global recovery module}.

\paragraph{Motion Verification} 
Given a leaf action $a=\omega(o_{1:n_\omega})$, the mapped skill $\pi$, and the current state of the environment $x_t$, we first verify that the preconditions of $\omega$ hold in the environment, such that $\text{pre}(\omega) \subseteq c_{\underline{\Psi}}(x_t)$. We then execute $\pi$ in the environment, returning the next state $x_{t+1} = f(x_t, \pi(o_{1:n_\pi}))$. We lastly verify that the effects $\text{eff}(\omega)$ equal those observed in the environment, such that $\text{eff}(\omega) = \text{eff}(c_{\underline\Psi}(x_i), c_{\underline\Psi}(x_{i+1}))$ (see Eq. \ref{eq:effect-formula}).
We verify that \textit{state-based} predicates hold in the simulation. \textit{State-independent} predicates define relations that do not change during planning and cannot be grounded from the low-level environment state, e.g. \textit{openable}.
We have to rely on the LLM to detect incorrectly defined \textit{state-independent} predicates, as we can not verify them in simulation.
%Consequently, %\textit{state-independent} cannot be verified from the low-level state. Instead, we rely on the \ac{llm} to detect incorrectly defined \textit{state-independent} predicates.
% \todo{why do we rely on the llm}

\paragraph{Global Recovery Module} 
A failed verification for an action $a_i$ in a plan $a_{1:k}$ can be caused by violated preconditions, misaligned effects, or an incorrect mapping to a predefined skill.
We propose a \textit{global recovery module} $\text{LLM}_{\text{reasoner}}$ that analyses the occurred verification failure to determine \textit{how to adapt the domain model}. $\text{LLM}_\text{reasoner}$ oversees the past model learning, with access to the  history of previous \ac{llm} interactions, and a summary of the motion verification failure, either showing the state misalignment, or the occurred exception raised while executing the skill.
$\text{LLM}_\text{reasoner}$ then determines the cause for the misalignment in the past model learning and suggests \textit{where} to adapt the domain model and \textit{how} to continue learning by outputing the operator $\omega$ that should be adapted.
%To decide \textit{how} to continue learning, $\text{LLM}_\text{reasoner}$ outputs the operator $\omega$ that should be adapted. 
Once decided, we retract to the affected level and re-prompt the related module to incorporate the fix.

\section{Evaluation}
We evaluate LODGE on three domains to answer (1) How well does LODGE learn operators with little predefined information and few environment interactions (Sec. \ref{subsec:planning-benchmark})? (2) How well can LODGE learn entire domains (predicates, predicate classifiers, and operators) (Sec. \ref{subsec:furniture-bench})? (3) Can LODGE optimize predicate classifiers from pseudo-labels (Sec. \ref{subsec:furniture-bench})?
% We focus on operator learning in Section \ref{subsec:planning-benchmark} and evaluate LODGE on learning entire domain models for a robotic assembly domain in Section \ref{subsec:furniture-bench}.

\subsection{Learning Operators for IPC Domains}
\label{subsec:planning-benchmark}
We evaluate operator learning with LODGE on the \textit{household} and \textit{logistics}  IPC domains.
% Tasks in both domains are long-horizon and household contains 22 skills, while logistics contains 6 skills (see \citet{guan_leveraging_2023} for details).
Since these domains operate on discrete state spaces, we do not learn predicate classifiers but ground the planner's state in the environment. Additionally, we focus on learning operators and assume the \textit{state-based} predicates are known.
LODGE learns operators for the first task of a domain and then refines it for each following task.
We limit environment interactions to 10 and re-planning iterations (e.g. to fix syntax errors) to 20 per task to demonstrate LODGE's efficiency in learning operators with few interactions.
We use the instructions and domain descriptions from \citet{guan_leveraging_2023}, but leave out their skill annotations, as they give a detailed description about the skills and are generally unavailable (see App. \ref{app:ipc-domains} for a exemplary skill).
% An exemplary skill with an annotation from the \textit{household} domain is the following:
% \begin{lstlisting}
% def heat_food_with_pan(food: str, pan: str):
%   "This action enables the robot to heat food which is heatable with a pan. The food should be placed on the pan, and the pan needs to be placed on a stove burner before executing this action. Note that the food is no longer pickupable after it has been heated."
% \end{lstlisting}
% \label{skill-signature}
\input{eval/domain-correctness-table}

We compare LODGE with \citet{guan_leveraging_2023} (without providing human feedback) and \citet{mahdavi_leveraging_2024}.
We additionally evaluated \citet{wong_learning_2023}'s method for domain learning, but it did not succeed in generating domain models for any of the two IPC domains (see Appendix \ref{app:wong-learning}). 
%We compare our method to Guan's domain construction method without correcting the domains with human feedback, as we are concerned about automatic domain learning.
For each learned domain model, we evaluate task success and estimate similarity to a predefined domain model using an adapted \ac{ew} metric~\citep{mahdavi_leveraging_2024} (see Appendix \ref{app:generalized-exploration-walk}). Each method is evaluated three times per domain, and results are averaged.
The \ac{ew} metric is defined as the harmonic mean of the two percentages that indicate how many plans randomly sampled from one domain are executable in the other domain. A value of 1.0 indicates that both domains are identical.
We compare all methods on GPT 4.1 mini and Llama 4 Maverick, and additionally evaluate the smaller Llama 4 Scout on the \textit{logistics} domain.

Table \ref{tab:domain-model-eval} evaluates the two IPC domains. While all methods perform well on \textit{logistics} with GPT 4.1 mini, LODGE remains robust with weaker \acp{llm} like Llama 4 Scout, whereas \citet{mahdavi_leveraging_2024} struggles without skill annotations. 
% This suggests that LODGE is more robust to the \acp{llm} intelligence than its baselines.
Moreover, for the \textit{household} domain, LODGE produces domain models closer to the predefined domain model, solving up to 44 \% of the tasks, while the domains of other approaches rarely yield successful plans.
% In contrast to LODGE, 
\citet{guan_leveraging_2023} and \citet{mahdavi_leveraging_2024} by default use skill annotations for domain learning. To assess the relevance of these annotations to their methods, we evaluate \citet{mahdavi_leveraging_2024} without passing the annotations (see Table \ref{tab:domain-model-eval} under `Mah. w/o ann.'). As \citet{guan_leveraging_2023} solely rely on skill annotations, we omit them from the analysis. Leaving out the annotations for \citet{mahdavi_leveraging_2024} decreases the domain model correctness for both \ac{ipc} domains significantly. Especially for the more complex \textit{household} domain, \citet{mahdavi_leveraging_2024} relies on the annotations and struggles to generate an accurate domain model without them.
% While all approaches reach good performance on the \textit{logistics} domain, LODGE makes the least assumptions. \citet{mahdavi_leveraging_2024} requires the operator signature (i.e. that \textit{drive-truck} has parameters \textit{?trk - truck ?from - location ?to - location ?city - city}) to compute the \ac{ew} metric of generated domain candidates and the executability of plans, effectively requiring a partially known ground-truth domain.

\subsection{Learning Domain Models for Robotics Domains}
\label{subsec:furniture-bench}
% Robotics domains differ fundamentally from \ac{ipc} domains. First, robot environments are inherently continuous, making state representation and reasoning more complex. Second, the executability of skills differs markedly: in \ac{ipc} domains, actions are only defined in states where their preconditions are satisfied (e.g., loading a package into a truck is only possible if both are at the same location).
% Lastly, such \textit{PDDL domains} are designed for classical planning, e.g. by defining separate skills for grasping from furniture versus grasping from receptacles.
% In contrast, robotic skills such as \textit{close-gripper} or \textit{move-up} can usually be executed in any state, but their outcome may include unintended effects, such as collisions. Consequently, the true executability of skills is not generally known a priori.
Robotics domains differ from \ac{ipc} domains by being continuous with generally executable but uncertain skills.
In this experiment, we learn entire domain models with LODGE, comprising predicates, predicate classifiers, and operators. 
LODGE only receives one predicate `\textit{assembled(?obj1, ?obj2)}' in the initial predicates $\Psi_\text{init}$, whose classifier indicates whether two parts have been assembled, as defined by the FurnitureBench benchmark~\citep{heo_furniturebench_2023}.
Since the planner's state and the continuous state of the environment are independent, we additionally construct and refine predicate classifiers for newly invented \textit{state-based} predicates (see Sec. \ref{sec:learning-predicates}). 
% These classifiers ground the symbolic state in the continuous environment state, which LODGE uses to detect and address misalignments within the \textit{recovery module} (see Sec. \ref{sec:motion-verification}).
We evaluate domain model learning as in Section \ref{subsec:planning-benchmark} and compare our algorithm to learn predicate classifiers to related methods.
% We evaluate LODGE on the \textit{lamp} assembly task of the FurnitureBench Benchmark \cite{heo_furniturebench_2023}. The skills in the FurnitureBench domain are general-purpose low-level robot controllers. We evaluate all methods for domain learning with GPT 4.1 mini as \ac{llm} and use GPT 4.1 as VLM. We additionally compare the algorithm we propose for learning predicate classifiers to related methods.

\paragraph{Domain Model Evaluation}
\input{eval/fb-pred-grounder-eval.tex}
Table \ref{tab:domain-model-eval} evaluates the learned domain model of the FurnitureBench domain and task success in assembling the \textit{lamp}.
While \citet{mahdavi_leveraging_2024} significantly outperforms \citet{guan_leveraging_2023} in IPC domains, it struggles in FurnitureBench.
The main cause lies in the EW metric, which, although smooth, decreases exponentially with the number of incorrect terms.
As a result, correcting individual errors for large (\textit{household}) or complex (\textit{FurnitureBench}) domains often has no effect on the score.
Moreover, the language feedback in \citet{mahdavi_leveraging_2024} is restricted to syntax errors and does not identify which operators or plans are problematic, which makes improvements difficult.
LODGE addresses these limitations by proposing a targeted adaptation of operators based on the predicate classifiers' feedback (see App. \ref{app:lodge-prompt}).
As a result, LODGE assembles the lamp in two of three seeds fully and achieves an average of 81 \% of the lamp assembly.

\begin{figure*}
    \centering
    \includesvg[width=\textwidth]{images/pddl-experiment.svg}
    \caption{The hierarchical planning of assembling the \textit{lamp} with LODGE for the FurnitureBench environment, including decomposition of \textit{pick-up(bulb)} and re-planning within \textit{screw-in(bulb)}.}
    \label{fig:furniture-bench}
\end{figure*}

Figure \ref{fig:furniture-bench} depicts the hierarchical domain learning of assembling \textit{bulb}, \textit{base} and \textit{hood} to a \textit{lamp}. The decomposition significantly reduces the plan length at the topmost level. LODGE invents predicates at the top level to capture high-level object relations, like \textit{aligned} or \textit{holding}. During decomposition, LODGE invents fine-grained predicates to track the state as needed, i.e. \textit{gripper-around-part} for the decomposition of \textit{pick-up} or \textit{touching} for the decomposition of \textit{screw-in}.
Additionally, LODGE reuses the decomposed domain model of \textit{pick-up(bulb)} for \textit{pick-up(hood)}.
\ifnotanonym
    A video of the planning, including the re-planning until a valid plan was found, is available on the website of the paper.
\fi

\paragraph{Predicate Classifier Evaluation}
While learning the hierarchical domain model, LODGE invents predicates and learns classifiers. We compare our method of learning classifiers to InterPreT \citep{han_interpret_2024} and Pix2Pred \citep{athalye_pixels_2025}.
% While Pix2Pred grounds predicates by handing an annotated image of the environment to a VLM, InterPreT generates a Python function for every predicate that implements its classifier.
Table \ref{tab:pred-grounding} shows the evaluation of all methods on the states observed while assembling the \textit{lamp}, evaluated on the F1 score $\mathcal{F}$ metric of Section \ref{sec:learning-predicates}.
\input{eval/plan-success-rate}
The VLM in \citet{athalye_pixels_2025} generates more accurate predicates than writing a Python classifier. We found that the Python classifier for some predicates yields a significantly lower F1 score than on others. This is either caused by an incorrect Python implementation (e.g. too restrictive checks such as comparing part rotations), coding errors, or unsuitable tolerances (e.g. \textit{on-table} uses a 1 mm distance tolerance, while perception noise is on the order of 1 cm).
In comparison, our method learns accurate Python classifiers by refining them with pseudo-labels.

\subsection{Ablation}
% We now present several ablations on the design choices of LODGE.
% The ablation highlights the importance of LODGE's components. We begin by comparing domain learning to refining plans on the environment.
% LODGE combines \acp{llm} with predicate classifiers. Subsequent ablations evaluate the effect of removing \acp{llm}-based operator learning or omitting grounding via predicate classifiers. Finally, we assess LODGE's robustness to variations in skill set size.

% \paragraph{How does domain learning compare to directly refining plans on the environment}
\paragraph{How does domain learning compare to plan refinement?}
Instead of learning domain models and thereby learning \textit{how to plan}, we can adapt the plan directly on the environment. %, e.g. using self-refinement with an \ac{llm}.
Table \ref{tab:abl-adapt-domain-vs-plan} compares the plan success of LODGE against an \ac{llm} planner \citep{valmeekam_planning_2023} that directly refines the plan, with and without access to skill annotations. LODGE solves more tasks with the same amount of environment interactions (\# Iters.) while jointly learning a domain model, which prevents future environment interactions to solve similar tasks.

\paragraph{How does LODGE compare to direct operator learning?}

Having invented predicates and predicate classifiers, we can construct operators with classical algorithms \citep{silver2021learning} on the interaction dataset $\mathcal{D}$ instead of using LODGE's \ac{llm}-based method for operator learning.
Therefore, we evaluate how LODGE compares to classical methods for operator learning on the same data.
\input{eval/ablation-cluster-intersect.tex}
Table \ref{tab:abl-cluster-intersect} compares the approaches of \citet{guan_leveraging_2023}, \citet{mahdavi_leveraging_2024} and LODGE with the \ac{cands} algorithm proposed by \citet{silver2021learning}.
We evaluate \ac{cands} on the same interactions LODGE sees during domain learning. 
\textit{\ac{cands} w. Py} uses labels generated by the correct predicate classifiers, while \textit{\ac{cands} w. VLM} uses Pix2Pred \citep{athalye_pixels_2025} for labeling, which also contains misclassifications.
The Table shows that operator learning with classical algorithms on limited amount of data reaches a low performance. Our method of combining \acp{llm} and predicate classifiers for grounding significantly outperforms directly inferring operators for a small dataset (20 transitions).

% \paragraph{Demonstrate how Hierarchical Planning improves Planning Speed}
% \include{eval/ablation-ai-planner}
% \include{eval/ablation-retaining-domains}

% \paragraph{Does predicate learning and grounding help to learn more accurate domain models?}
\input{eval/ablation-no-grounder.tex}
\paragraph{Does predicate grounding improve domain models?}
We evaluate LODGE without learning predicate classifiers in the FurnitureBench  \textit{lamp} assembly domain. Table \ref{tab:abl-grounding-helps} shows that not grounding the planners' state in the environment significantly impacts domain model correctness and planning success, underlining the importance of grounding domains in the environment.

% \paragraph{Show that LODGE is robust against skill set size}
\paragraph{Is LODGE robust to the skill set size?}
LODGE constructs operators only for skills needed to solve the current task, and hierarchical abstraction reduces the number of operators at each level. We evaluate the effect of skill set size, by increasing the number of skills from 22 to 72 in the \textit{household} domain. Despite the threefold increase in available skills, the number of learned operators did not increase, and the use of input and output token increased only slightly by 44\% and 25\%, respectively.

\section{Conclusion}
\label{sec:conclusion}
We propose LODGE, a framework that uses \acp{llm} and environment grounding to learn hierarchical domain models by iteratively refining an abstract domain into subdomains. We propose to learn classifiers for newly invented predicates and refine them on pseudo-labeled environment interactions.
A novel \textit{global recovery module} analyzes misalignments between the domain model and environment by evaluating the classifiers and guides realigning the domain with the environment.
Evaluation on three domains shows that LODGE learns more accurate domain models that yield higher task success than existing methods.
On the robotics domain, we demonstrate that LODGE can refine domain models from learned classifiers for continuous environments that are not tailored to formal languages.

While LODGE shows strong performance on hierarchical domain learning, several directions remain for future work.
First, domain decomposition requires aligning hierarchy levels, which is non-trivial due to goal overshoots and side effects. Future work could investigate on more robust approaches for smaller \acp{llm}.
Second, our robotics evaluation is limited to the FurnitureBench domain, as no benchmarks exist for domain learning on continuous environments. 
Existing robotics benchmarks either target RL or operate on discrete spaces \citep{silver_pddlgym_2020}.
% Addressing this gap will require developing suitable benchmarks.
Third, constructing large domain models with LODGE from scratch is slow. More sophisticated decomposition could improve learning speed, e.g. by sharing operators across hierarchy levels.
Finally, combining LODGE with efficient, constraint-based path planners could realize highly flexible \ac{tamp} systems % that jointly optimize robot programs at the task and motion level
\citep{silver2021learning}, which requires further research into representations bridging PDDL-style planners, constraint-based motion planning and \ac{llm} reasoning.
\ifnotanonym
    \section{Acknowledgement}
    This  work  was  supported  by  the  German  Federal  Ministry  of  Education and Research (project RobInTime, grant 01IS25002B).
\fi

\bibliographystyle{IEEEtranN}
\bibliography{bibliography_config,bibliography}

%%%%%%%%%%%%%%%%%%%%%%%%%%%%%%%%%%%%%%%%%%%%%%%%%%%%%%%%%%%%

\include{appendix}

% \newpage
% \include{neurips-checklist}

\end{document}

%% file: eval/domain-correctness-table.tex
\begin{table*}[]
    \centering
    \begin{tabular}{ll cccc}
        \toprule
        Env & Model         & Guan        & Mah. w/ ann. & Mah. w/o ann. & LODGE (ours)                     \\ \midrule
        \multirow{3}*{Logistics}
            & Llama 4 Scout & 0.09 / 0.05 & 0.97 / 0.95   & 0.70 / 0.37 & 0.92 / 0.87 \\
            & Llama 4 Mav.  & 0.08 / 0.05 & 1.00 / 1.00   & 1.00 / 1.00 & 1.00 / 1.00 \\
            & GPT 4.1 mini  & 0.90 / 0.68 & 1.00 / 1.00   & 0.99 / 0.98 & 1.00 / 1.00 \\
            \cmidrule{2-6}
        \multirow{3}*{Household}
            & Llama 4 Mav.  & 0.03 / 0.00 & 0.38 / 0.00   & 0.07 / 0.00 & 0.57 / 0.25 \\
            & GPT 4.1 mini  & 0.06 / 0.00 & 0.42 / 0.00   & 0.20 / 0.03 & 0.69 / 0.44 \\
            \cmidrule{2-6}
       Furniture
            & GPT 4.1 mini  & 0.26 / 0.06 & 0.30 / 0.25   & 0.23 / 0.17 & 0.71 / 0.81 \\
        \bottomrule
    \end{tabular}
    \caption{Avg@3 (Exploration Walk/Tasks solved) for three domains}
    \label{tab:domain-model-eval}
\end{table*}

%% file: eval/fb-pred-grounder-eval.tex
\begin{wraptable}[9]{r}{0pt}
    \begin{tabular}{lc}
        \toprule
        Method         & F1 Score \\
        \midrule
        InterPreT (Py) & 0.71     \\
        Pix2Pred (VLM) & 0.84     \\
        LODGE (ours)   & 1.00     \\
        \bottomrule
    \end{tabular}
    \caption{Predicate Grounding}
    \label{tab:pred-grounding}
\end{wraptable}

%% file: eval/plan-success-rate.tex
\newcommand{\outof}[2]{#1{\tiny\ of #2}}

\begin{wraptable}[14]{r}{0pt}
    \begin{tabular}{lc rr}
        \toprule
        Planner Type
         & \# It. & LOG            & HH             \\ \midrule
        \multirow{2}*{\makecell{LLM Planner         \\Ann.=yes}}
         & 0      & \outof{1}{21}  & \outof{3}{24}  \\
         & 10     & \outof{18}{21} & \outof{11}{24} \\
        %  \cmidrule{2-4}
        \multirow{3}*{\makecell{LLM Planner         \\Ann.=no}}
         & 0      & \outof{1}{21}  & \outof{1}{24}  \\
         & 10     & \outof{13}{21} & \outof{6}{24}  \\
         & 20     & \outof{13}{21} & \outof{13}{24} \\
        %  \cmidrule{2-4}
        LODGE (ours)
         & 10     & \outof{21}{21} & \outof{11}{24} \\ 
        \bottomrule
    \end{tabular}
    \caption{Task success in Logistics (LOG) and Household (HH) domains, comparing domain learning with refining plans via environment interactions (\#It.) and skill annotations (Ann.).}
    \label{tab:abl-adapt-domain-vs-plan}
\end{wraptable}

%% file: eval/ablation-cluster-intersect.tex
\begin{wraptable}[12]{r}{0pt}
    \centering
    \begin{tabular}{lc}
        \toprule
        Method             & Domain Eval. \\ \midrule
        Guan               & 0.26 / 0.06 \\
        Mahdavi w/o Annot. & 0.23 / 0.17 \\
        LODGE (ours)       & 0.71 / 0.78 \\ \midrule
        \acs{cands} w. Py         & 0.23 / 0.11 \\
        \acs{cands} w. VLM        & 0.17 / 0.00 \\
        \bottomrule
    \end{tabular}
    \caption{Fitting operators directly on the environment interactions}
    \label{tab:abl-cluster-intersect}
\end{wraptable}

%% file: eval/ablation-no-grounder.tex
\begin{wraptable}[9]{r}{0pt}
    \begin{tabular}{lc}
        \toprule
        Grounding & EW / Succ.  \\ \midrule
        w/        & 0.78 / 0.71 \\
        w/o       & 0.38 / 0.14 \\
        \bottomrule
    \end{tabular}
    \caption{Importance of predicate grounding for LODGE on \textit{FurnitureBench}}
    \label{tab:abl-grounding-helps}
\end{wraptable}%

%% file: appendix.tex
\appendix

\section{Hierarchical Domain Learning}
\input{eval/alg-decomposition}
Algorithm \ref{alg:decomposition} shows the learning of a hierarchical domain model.

\section{Error Reasoner}
The central error reasoner can invoke corrections at any hierarchy level. We alternatively tested retracting one hierarchy level at a time to self-verify each previously instructed \ac{llm} given the observed simulation error. However, the \acp{llm} tend to correct themselves even if they cause for the misalignment lies somewhere else. Additionally, a level-by-level retracting is more computationally expensive with increasing number of hierarchy levels, making it unprobable to retract multiple layers, while the central reasoner can skip any hierarchy levels ad trigger re-planning.

To further enhance coherence, we introduce an in-place modification mechanism for \ac{llm} messages when traversing within a hierarchy level. This mechanism summarizes information or results from lower levels and injects them as a single, refined message, as if the \ac{llm} had directly produced the correct output without enlargening the chat history. For example, during decomposition, the \ac{llm} initially proposes a domain, which is iteratively refined to correct syntax, semantics, or planning infeasibility. The final, corrected domain is then re-injected into the original response, keeping the chat history compact and concealing intermediate errors.

\section{VLM Grounding}
\label{app:vlm-grounding}
We obtain $s'_i$ from a VLM (as in \citet{athalye_pixels_2025}) given the transition $t_i$ and an image of the scene at state $x'_i$. The start state $x_0$ is labeled directly from $x_0$ and the image.

To reduce the number of samples to ground with the VLM, we compare the similarity of a new state with previously seen states. Only significantly new states are grounded by the VLM. A state that is similar to an existing state uses the same VLM grounding. The intuition for this is that the image of similar states are very similar. To estimate similarity, we compute the equivalence class for every sample by
\begin{align*}
    \text{neigh}(\pi, x) = |\{(x_j, \pi_j, s_j) \in \mathcal{D} : \pi_j = \pi \wedge \|x_j - x\| < \tau \}|.
\end{align*}
Only the first sample of this equivalence class is grounded by the VLM. All other samples in that class copy that grounding.
To ground the states with the VLM, we use the same prompts as \citet{athalye_pixels_2025}. We additionally pass known perception information and predicate descriptions to the VLM, e.g.
\begin{lstlisting}
Pose estimation of all objects and the robot gripper (3D poses are in meters):
arm
- gripper_center: [0.567, 0.055, 0.124]
- gripper_closed: False
table
- surface_z: -0.016
lamp_base
- center: [0.455, 0.050, 0.017]
- orientation: [3.140, -0.000, -1.380]
lamp_bulb
- center: [0.433, 0.195, 0.015]
- orientation: [-3.140, 1.420, 0.040]
lamp_hood
- center: [0.459, -0.110, 0.035]
- orientation: [3.140, -0.000, 1.540]
Predicates:
# Grounding assembled: Evaluates whether obj1 has been assembled with obj2 the way they should be, e.g. when obj1 has been screwed into obj2, or obj1 has been put on top of obj2.
assembled(lamp_base, lamp_bulb)
...
\end{lstlisting}

\section{Python Predicate Classifier}
\label{app:py-predicate-grounding}
All Python classifiers can use following perception API. 
\begin{lstlisting}[language=Python]
from dataclasses import dataclass
from typing import Tuple
import math

@dataclass
class Part:
    bounding_box: Tuple[float, float, float, float, float, float]  # (x_min, y_min, z_min, x_max, y_max, z_max)
    center: Tuple[float, float, float]  # (x, y, z)
    orientation: Tuple[float, float, float]  # (roll, pitch, yaw) in radian

@dataclass
class Table:
    surface_z: float

@dataclass
class Robot:
    gripper_center: Tuple[float, float, float]  # (x, y, z) of midpoint between jaws
    gripper_closed: bool

def angle_diff(a: float, b: float):
    """Compute the minimal difference between two angles in radians."""
    return ((a - b) + math.pi) % (2 * math.pi) - math.pi

def get_robot() -> Robot: ...
def get_table() -> Table: ...
def get_all_parts() -> List[Part]: ...
\end{lstlisting}
Additionally, a predicate classifier can call other already invented classifiers. 
\subsection{Generate Python classifier}
The prompt to the \ac{llm} to generate a predicate classifier for \textit{aligned(?obj1, ?obj2)} is 
\begin{lstlisting}
Domain Knowledge:
{domain description}

Given Python code:
```python
{perception api}

def holding(r: Robot, p: Part) -> bool:
    """The robot is currently holding the part with its gripper.
    Returns: True if the predicate holds, False otherwise.""" ...
def on_table(p: Part, t: Table) -> bool:
    """The part is resting on the table.
    Returns: True if the predicate holds, False otherwise.""" ...
```

```python
def aligned(p1: Part, p2: Part) -> bool:
    """The held part is aligned with the fixed part for assembly (xy position and roll/pitch orientation aligned).
    Returns: True if the predicate holds, False otherwise.""" ...
```
Predicate: (aligned ?p1 - part ?p2 - part): The held part is aligned with the fixed part for assembly (xy position and roll/pitch orientation aligned).

Can you add the python body that implements the grounding of this predicate?
You must define hyperparameters or constants as kwargs with default values.
Where possible, use the already defined python functions to reduce code duplication.
Your response should contain two sections
[START OUTLINE]
# Requirements to Check
[list all requirements the function must validate. Only include checks that can be performed with the provided context (skip/ignore others).]
# Predicate Grounding
[insert the python code enclosed in ```python ```. If you think the predicate cannot be grounded reliably with the provided context, return `none` and no python code. Don't define methods you do not implement or rely on information not provided.]
# Grounder Description
[insert a description what the grounder function tests for. It should be short but complete.]
[END OUTLINE]
\end{lstlisting}
The prompt contains python functions from already invented classifiers \textit{holding}, \textit{on\_table}, and \textit{aligned}. Also, it can be seen that the signature of the Python functions mimics the predicate signature.

The \ac{llm} then defines the function body of the predicate classifier in its response and adds required hyperparameters as additional keyword arguments to the function signature.
\begin{lstlisting}
[START OUTLINE]  
# Requirements to Check  
- The xy positions of p1 and p2 are within a given positional tolerance (`pos_tol`).  
- The roll angles of p1 and p2 differ by no more than a given angular tolerance (`angle_tol`).  
- The pitch angles of p1 and p2 differ by no more than a given angular tolerance (`angle_tol`).  
- The yaw difference is ignored.  

# Predicate Grounding  
```python
def aligned(p1: Part, p2: Part, pos_tol: float = 0.01, angle_tol: float = 0.1) -> bool:
    """The held part is aligned with the fixed part for assembly (xy position and roll/pitch orientation aligned).

    Args:
        p1: The held part.
        p2: The fixed part.
        pos_tol: Positional tolerance in meters for xy alignment.
        angle_tol: Angular tolerance in radians for roll and pitch alignment.

    Returns:
        True if aligned, False otherwise.
    """
    # Check xy position closeness
    dx = p1.center[0] - p2.center[0]
    dy = p1.center[1] - p2.center[1]
    pos_close = (dx * dx + dy * dy) <= (pos_tol * pos_tol)

    # Check roll difference
    roll_diff = abs(angle_diff(p1.orientation[0], p2.orientation[0]))
    # Check pitch difference
    pitch_diff = abs(angle_diff(p1.orientation[1], p2.orientation[1]))

    return pos_close and roll_diff <= angle_tol and pitch_diff <= angle_tol
```  
# Grounder Description  
The grounder checks whether two parts are aligned by verifying that their xy positions are within a specified positional tolerance and that their roll and pitch angles differ no more than a given angular tolerance. The yaw angle difference is ignored since it is not required for alignment.  
[END OUTLINE] 
\end{lstlisting}

\subsection{Refine Python classifier}
The prompt to refine Python classifiers contains the current implementation, as well as up to three samples from the interaction dataset $\mathcal{D}$. It additionally provides the perception information for every samples, as well as the classifier's output and the pseudo-label.
\begin{lstlisting}
{domain description}

{code api}

{current predicate classifier}

This is a function that grounds the PDDL predicate. However, we evaluated it against a dataset labeled by an VLM and found following errors:
1.
- VLM predicates: (gripper_around lamp_bulb)
- Grounder predicates: (not (gripper_around lamp_bulb))
- Referenced Predicate Evals:

- Variables:
    -arm: Robot(gripper_center=[0.456, 0.183, 0.014], gripper_closed=False)
    -table: Table(surface_z=-0.016)
    -lamp_base: Part(bounding_box=[0.405, 0.002, -0.015, 0.505, 0.101, 0.048], center=[0.455, 0.052, 0.017], orientation=[3.14, -0.0, -1.38])
    -lamp_bulb: Part(bounding_box=[0.37, 0.149, -0.015, 0.49, 0.209, 0.045], center=[0.43, 0.179, 0.015], orientation=[-1.75, 0.54, 1.5])
    -lamp_hood: Part(bounding_box=[0.415, -0.154, -0.015, 0.503, -0.066, 0.085], center=[0.459, -0.11, 0.035], orientation=[-3.14, -0.0, 1.54])

    Your response should contain three sections
[START OUTLINE]
# Error Analysis
[evaluate the errors by tracing the values and result in the code.]
# Suggested Fixes
[list all checks in the code that contributed to the error. For every of them, verify they are required given information about the predicate. Consider also new checks that should be implemented.]
# Fixed Code
[insert your suggested changes and output the updated Python code. The fixes should correct the function to fix as many errors as possible while not violating the predicate definition and description. Do not add methods with a placeholder body or `NotImplementedError`.]
# Grounder Description
[insert a description what the grounder function tests for. It should be short but complete.]
[END OUTLINE]
\end{lstlisting}

Based on that, the \ac{llm} analyzes the cause for the different classifications of VLM and Python classifier and suggests adaptations to the Python classifier, followed by an adapted implementation of the function body.
\begin{lstlisting}
[START OUTLINE]  
# Error Analysis  
The reported error example shows that the VLM considers `(gripper_around lamp_bulb)` true, but the current grounding function returns false. Analyzing the code, the failure likely arises from the orientation check, because:  
- The gripper orientation roll and pitch are hardcoded as zero.  
- The lamp_bulb has roll = -1.75 rad (~ -100deg), pitch = 0.54 rad (~31deg), both far from zero.  
- Orientation difference tolerance is 0.3 rad (~17deg), so this difference fails the orientation check, causing the predicate to return False.  

Other checks:  
- Gripper closed is False, so pass.  
- Distance between gripper and part xy centers looks small enough (gripper at [0.456, 0.183], bulb at [0.43, 0.179]) => dx=0.026, dy=0.004 => dist_xy ~0.026, within 0.05 tolerance.  
- Vertical difference dz = abs(0.014 - 0.045) = 0.031, < 0.05, passes.  

Conclusion: orientation check fails due to assumed gripper orientation zero, but robot orientation isn't known, so the check is too strict or invalid with current data.  

# Suggested Fixes  
- Remove or relax orientation checks because gripper orientation data is unknown / not available. This aligns with the comment in code: "If this assumption is too strong, this step can be omitted or weakened."  
- Keep other checks intact (gripper open, xy distance, vertical height relative to part top).  
- Optionally, allow a larger orientation tolerance or skip orientation check completely for roll and pitch because gripper orientation is not provided and robot does not restrict roll/pitch strictly in grasping parts.  
- Possibly rename variable to clarify that gripper orientation is assumed unknown and that orientation check is being skipped.  
- Keep all distance and gripper open checks mandatory to maintain semantic correctness.  

# Fixed Code  
```python
def gripper_around(p: Part, xy_tolerance: float=0.05, z_tolerance: float=0.05) -> bool:
    """
    True if the robot's gripper is positioned around and ready to grasp part p.

    Args:
        p (Part): The part to check against
        xy_tolerance (float): Max allowed horizontal xy distance (m)
        z_tolerance (float): Max allowed vertical distance (m) from gripper to part top surface

    Returns:
        bool: True if gripper is horizontally close, vertically near the top of the part, and open.
    """
    robot = get_robot()
    # Gripper must be open (ready to grasp)
    if robot.gripper_closed:
        return False

    # Horizontal distance in xy between gripper center and part center
    dx = robot.gripper_center[0] - p.center[0]
    dy = robot.gripper_center[1] - p.center[1]
    dist_xy = math.sqrt(dx * dx + dy * dy)
    if dist_xy > xy_tolerance:
        return False

    # Vertical position check: gripper height close to part top (bounding_box z_max)
    part_top_z = p.bounding_box[5]
    dz = abs(robot.gripper_center[2] - part_top_z)
    if dz > z_tolerance:
        return False

    # Orientation check omitted due to unknown gripper orientation

    return True
```

# Grounder Description  
This grounding function checks whether the robot gripper is open and located approximately horizontally (within 5 cm) above the center of the part, and vertically near (within 5 cm) the top surface of the part, making it properly positioned "around" the part ready for grasping. The function assumes gripper orientation roll and pitch unknown and does not enforce orientation alignment.  
[END OUTLINE]    
\end{lstlisting}
We only retain the adapted Python classifier if it not yields a worse f1 score $\mathcal{F}$ on the interaction dataset, thereby mitigating hallucinations.

\section{Wong et al. - Learning domain models}
\label{app:wong-learning}
\citet{wong_learning_2023} propose to learn domain models by generating a large set of operator candidates. This set is then filtered by the ones that are used to solve a predefined set of planning problems. \citet{wong_learning_2023} redo the operator sampling two times, which we also used for evaluating it on the IPC domains. However, for both IPC domains, their method was not able to generate a set of operators that solved the planning for any of the tasks. Therefore, all operators were discarded, resulting in an empty set of operators. We used the official implementation from \citet{wong_learning_2023}.

\section{IPC Domain}
\label{app:ipc-domains}
In the following, we will list natural language instructions and sample plans for some tasks in the \textit{logistics} and \textit{household} domains.

An exemplary skill in the \textit{household} domain with annotations looks like following:
\begin{lstlisting}
def heat_food_with_pan(food: str, pan: str):
  "This action enables the robot to heat food which is heatable with a pan. The food should be placed on the pan, and the pan needs to be placed on a stove burner before executing this action. Note that the food is no longer pickupable after it has been heated."
\end{lstlisting}
\subsection{Logistics}

Task 1: Transport package \texttt{package\_0} to location \texttt{location\_2}
\begin{lstlisting}
load_truck('package_0', 'truck_1')
fly_plane('plane_0', 'location_1', 'location_0')
unload_truck('package_0', 'truck_1')
load_plane('package_0', 'plane_0')
fly_plane('plane_0', 'location_0', 'location_1')
unload_plane('package_0', 'plane_0')
drive_truck('truck_0', 'location_2', 'location_1')
load_truck('package_0', 'truck_0')
drive_truck('truck_0', 'location_1', 'location_2')
unload_truck('package_0', 'truck_0')
\end{lstlisting}

Task 2: Transport package \texttt{package\_3} to location \texttt{location\_3}, package \texttt{package\_4} to location \texttt{location\_0}, package \texttt{package\_1} to location \texttt{location\_2}, package \texttt{package\_0} to location \texttt{location\_0} and package \texttt{package\_2} to location \texttt{location\_4}
\begin{lstlisting}
load_truck('package_1', 'truck_1')
load_plane('package_4', 'plane_0')
load_plane('package_0', 'plane_0')
drive_truck('truck_1', 'location_4', 'location_5')
unload_truck('package_1', 'truck_1')
load_plane('package_1', 'plane_0')
fly_plane('plane_0', 'location_5', 'location_2')
unload_plane('package_4', 'plane_0')
load_truck('package_4', 'truck_0')
load_plane('package_3', 'plane_0')
unload_plane('package_1', 'plane_0')
unload_plane('package_0', 'plane_0')
load_truck('package_0', 'truck_0')
drive_truck('truck_0', 'location_2', 'location_0')
unload_truck('package_4', 'truck_0')
unload_truck('package_0', 'truck_0')
drive_truck('truck_0', 'location_0', 'location_1')
load_truck('package_2', 'truck_0')
drive_truck('truck_0', 'location_1', 'location_2')
unload_truck('package_2', 'truck_0')
load_plane('package_2', 'plane_0')
fly_plane('plane_0', 'location_2', 'location_5')
unload_plane('package_3', 'plane_0')
load_truck('package_3', 'truck_1')
unload_plane('package_2', 'plane_0')
load_truck('package_2', 'truck_1')
drive_truck('truck_1', 'location_5', 'location_3')
unload_truck('package_3', 'truck_1')
drive_truck('truck_1', 'location_3', 'location_4')
unload_truck('package_2', 'truck_1')
\end{lstlisting}

\subsection{Household}

Task 1: put \texttt{apple\_2} on \texttt{side\_table\_2}
\begin{lstlisting}
go_to("dining_table_1")
put_on("mug_1", "dining_table_1")
go_to("fridge_1")
open("fridge_1")
pick_up("lunch_box_2", "fridge_1")
go_to("dining_table_1")
put_on("lunch_box_2", "dining_table_1")
open_small("lunch_box_2")
pick_up_from("apple_2", "lunch_box_2")
go_to("side_table_2")
put_on("apple_2", "side_table_2")
\end{lstlisting}

Task 2: heat \texttt{pizza\_1} with \texttt{pan\_2}, and put it on \texttt{plate\_2}
\begin{lstlisting}
go_to("dining_table_1")
put_on("mug_1", "dining_table_1")
go_to("cabinet_2")
open("cabinet_2")
pick_up("pizza_box_1", "cabinet_2")
go_to("countertop_2")
put_on("pizza_box_1", "countertop_2")
open_small("pizza_box_1")
go_to("drawer_1")
pick_up("pan_2", "drawer_1")
go_to("countertop_2")
put_on("pan_2", "countertop_2")
transfer("pizza_1", "pizza_box_1", "pan_2")
pick_up("pan_2", "countertop_2")
go_to("stove_burner_2")
put_on("pan_2", "stove_burner_2")
heat("pizza_1", "pan_2")
go_to("cabinet_3")
open("cabinet_3")
pick_up("plate_2", "cabinet_3")
go_to("countertop_2")
put_on("plate_2", "countertop_2")
go_to("stove_burner_2")
pick_up("pan_2", "stove_burner_2")
go_to("countertop_2")
put_on("pan_2", "countertop_2")
transfer("pizza_1", "pan_2", "plate_2")
\end{lstlisting}

\section{Generalized Exploration Walk}
\label{app:generalized-exploration-walk}
We evaluate the domain models by computing an adapted version of the EW metric of \citet{mahdavi_leveraging_2024}. It contains two adaptations: 
\paragraph{(1) Evaluating context-centric domain models with the EW scores.} 
Since LODGE learns not a full domain model, but context-centric domain models, the final domain model does not contain operators for all aspects of the domain. For example for setting a table, LODGE would not learn operators for wiping the floor. Consequently, we have to adapt the EW score do evaluate context-centric domain models. For every task, we bootstrap the generated and predefined domain model to only contain operators required to solve that task. We evaluate the EW score on this task-centric domain model. We do this for all tasks and average the metrics.
\paragraph{(2) Sampling over operators instead of actions} 
When sampling a random walk from a domain model, \citet{mahdavi_leveraging_2024} sample the next action uniformly from all available actions. With this, operators that can be executed with many different objects dominate the EW score. For example for the \textit{household} domain, the agent can always \textit{go-to} another furniture. The sampled random walk consequently contains mostly \textit{go-to} actions and only rarely evaluates the similarity of other operators.
To fix this, we sample uniformly over the applicable operators and then sample uniformly over all actions from the selected operator. With that, all operators are evaluated evenly.

Additionally, we sample 500 plans compared to 100 used by \citet{mahdavi_leveraging_2024}.

\section{LODGE Prompts}
\label{app:lodge-prompt}

\subsection{$\text{LLM}_\text{domain}$}
System
\begin{lstlisting}
You are a planning expert tasked with developing a world model for planning based on a user instruction.

--- 

<!-- when generating content for the sections listed below, follow the specified format exactly. You can leaf out sections you don't need. -->
### Explanation
<!-- task specific explanation and chain-of-thought reasoning -->

### Change/Add Action(s)
1. {action-name-1}: {(add|edit|delete)} 
    - Description: {description what happens during the action}
    - PDDL Definition:
        ```pddl
        (:action {pddl_action_name}
            :parameters {pddl_action_parameters}
            :precondition {pddl_action_preconds}
            :effect {pddl_action_effects}
        )
        ```

### Change/Add Predicate Definitions
<!-- `state` predicates describe the current situation of objects in the world. Their description must specify the conditions under which they hold, so a classifier can later decide from perception data (e.g., 6D poses) whether the predicate is true.-->
- ({predicate_name} {predicate_args...}): {state|other}. {predicate_description}

### Change Initial State
<!-- add predicates that should be changed, without text, leave 'None' if no change -->
{predicate_4}: {true|false|remove}
{predicate_8}: {true|false|remove}

### Change Goal State
<!-- add predicates that should be changed, without text, leave 'None' if no change -->
{predicate_7}: {true|false|remove}
\end{lstlisting}
User
\begin{lstlisting}
### User Instruction
assemble the lamp

### Domain Knowledge
{domain description}

### Predicates
- (assembled ?obj1 - part ?obj2 - part). other: Evaluates whether obj1 has been assembled with obj2 the way they should be, e.g. when obj1 has been screwed into obj2, or obj1 has been put on top of obj2.

### Actions

### Types
robot table part - object

### Objects
arm - robot table - table lamp_base lamp_bulb lamp_hood - part

### Skill Library
- def hover_above_part(part: str):
- def set_gripper_around_part(part: str):
- def move_linear_up():
- def open_gripper():
- def close_gripper():
- def align_orientation_for_assembly(held_part: str, fixed_part: str):
- def screw_touching_parts_together(held_part: str, fixed_part: str):
- def move_linear_down_until_touching(held_part: str, fixed_part: str):

### Initial State
init state will be replaced

### Your Task
1. Define the goal: Based on the user instruction, create a PDDL goal that reflects the objective the user wants to accomplish.
2. List of predicates: Inspect the predicates that will be used to describe the state of the world and the relationships between entities in the domain.
3. Define actions: 
    - Based on the goal and available skills in the skill library, define a set of PDDL actions that enable planning toward the goal.
    - Preferably define high-level actions that abstract over one or more low-level skills to support hierarchical planning, but don't add object specific actions, e.g. `unstack-block-2`.
    - Each PDDL action should:
        - Have a clear and descriptive name
        - Include a general-purpose description (not instance-specific)
        - Include a PDDL definition: (:action <action_name> :parameters <parameters> :precondition <precondition> :effect <effect>)
    - You are encouraged to:
        - Define high-level composite actions when they simplify planning.
        - Define individual PDDL actions for each skill you intend to use that has no PDDL action yet.
        - If PDDL actions are already given in the user prompt, use them and only add new ones if necessary.
        - Avoid unnecessary actions - only include those essential to achieving the goal.
        - Don't create object specific actions, e.g. `switch-block-2-and-block-3`, but rather generate a general action `switch-blocks`.
        - Temporal ordering of subtasks can later be enforced by the LLM-planner
\end{lstlisting}
Assistant
\begin{lstlisting}
### Explanation
The goal is to assemble the lamp; (...)

### Change/Add Action(s)
1. grasp_part: add  
    - Description: Robot arm hovers above a part, grasps it by closing the gripper, and lifts it up.  
    - PDDL Definition:
    ```pddl
    (:action grasp_part
        :parameters (?r - robot ?p - part ?t - table)
        :precondition (and (not (assembled ?p ?_)) ; part is free to grasp (not already assembled, simplified)
                            )
        :effect (and (holding ?r ?p)
                        (not (on_table ?p ?t))
                        )
    )
    ```

(...)

4. assemble_parts: add  
    - Description: Assemble two parts by screwing them together while the robot holds one part aligned and touching the other.  
    - PDDL Definition:
    ```pddl
    (:action assemble_parts
        :parameters (?r - robot ?part1 - part ?part2 - part)
        :precondition (and (holding ?r ?part1)
                            (touching ?part1 ?part2)
                            )
        :effect (and (assembled ?part1 ?part2)
                        (not (holding ?r ?part1))
                        )
    )
    ```

### Change/Add Predicate Definitions
- (holding ?r - robot ?p - part): state. The robot is currently holding the part with its gripper.
- (on_table ?p - part ?t - table): state. The part is resting on the table.
- (aligned ?p1 - part ?p2 - part): state. The held part is aligned with the fixed part for assembly (xy position & roll/pitch orientation aligned).
- (touching ?p1 - part ?p2 - part): state. The held part is in contact with the fixed part, ready for assembly.

### Change Goal State
(assembled lamp_bulb lamp_base): true  
(assembled lamp_hood lamp_base): true
\end{lstlisting}

\subsection{$\text{LLM}_\text{translate}$}
User
\begin{lstlisting}
### Predefined Skills
```python
def hover_above_part(part: str): ...
def set_gripper_around_part(part: str): ...
def move_linear_up(): ...
def open_gripper(): ...
def close_gripper(): ...
def align_orientation_for_assembly(held_part: str, fixed_part: str): ...
def screw_touching_parts_together(held_part: str, fixed_part: str): ...
def move_linear_down_until_touching(held_part: str, fixed_part: str): ...
```

### Objects
arm - robot table - table lamp_base lamp_bulb lamp_hood - part

### Predicate Definitions
(assembled ?obj1 - part ?obj2 - part): (...)
(...)
(touching ?p1 - part ?p2 - part): (...)

### PDDL Action Definition
grasp-part: add
    - Description: Robot arm hovers above a part, grasps it by closing the gripper, and lifts it up.
    - PDDL Definition:
        ```pddl
        (:action grasp-part
            :parameters (?r - robot ?p - part ?t - table)
            :precondition (and (on_table ?p ?t) (not (holding ?r ?p)))
            :effect (and (holding ?r ?p) (not (on_table ?p ?t)))
        )        
        ```

### Task
The current state fulfills the PDDL action `grasp-part`'s preconditions. Your task is to propose a sequence of predefined skills to reach the effect of the PDDL action.

Here is an outline of what your response should look like:
[START OUTLINE]
# Outline Current State
[given the action preconditions, outline the current state]
# Action Description
[insert your analysis what the PDDL action is trying to achieve based on the current state and the effects that must be reached]
# Skill Mapping
[insert a bullet list of predefined skills as '<function_name>(<arg1>, <arg2>, ...)' that, starting from the current state, reach the effects. Don't include skills to `confirm` the preconditions. Use variables `from variables import r_robot, p_part, t_table` to reference pddl parameters, or strings if you need additional arguments where no pddl parameter exists]
[END OUTLINE]
\end{lstlisting}
Assistant
\begin{lstlisting}
[START OUTLINE]
# Outline Current State
- The robot arm (r_robot) is not holding the part (p_part).
- The part (p_part) is on the table (t_table).

# Action Description
- The grasp-part action aims for the robot arm to pick up the part from the table.
- The preconditions require the part to be on the table and not already held.
- The effect is that the robot will be holding the part and it will no longer be on the table.
- To achieve this, the robot needs to position its gripper over the part, close the gripper around the part, and then lift the part up from the table.

# Skill Mapping
- hover_above_part(p_part)
- set_gripper_around_part(p_part)
- close_gripper()
- move_linear_up()
[END OUTLINE]
\end{lstlisting}

\subsection{$\text{LLM}_\text{decompose}$}
User
\begin{lstlisting}
You are given a high-level PDDL action and supporting context. Your task is to decompose this high-level action into a set of meaningful, lower-level PDDL actions that result in the same effect.

### High-level Action `(grasp-part arm lamp_bulb table)`
grasp-part: add
    - Description: Robot arm hovers above a part, grasps it by closing the gripper, and lifts it up.
    - PDDL Definition:
        ```pddl
        (:action grasp-part
            :parameters (?r - robot ?p - part ?t - table)
            :precondition (and (on_table ?p ?t) (not (holding ?r ?p)))
            :effect (and (holding ?r ?p) (not (on_table ?p ?t)))
        )        
        ```

### Predicates
{predicates}

### Actions

### Types
robot table part - object

### Objects
arm - robot table - table lamp_base lamp_bulb lamp_hood - part

### Initial State
(on_table lamp_hood table) (on_table lamp_base table) (on_table lamp_bulb table)

### Goal State
(and (not (on_table lamp_bulb table)) (holding arm lamp_bulb))

### Skill Library
- def hover_above_part(part: str):
(...)

### Suggested Decomposition
```pddl
hover_above_part(p_part)
set_gripper_around_part(p_part)
close_gripper()
move_linear_up()
```

---

### Instructions
Follow the steps below to complete the decomposition:
1. Describe the Initial State
2. Understand the High-Level Action: Examine the :precondition and :effect. Identify what state change it induces.
3. Define Lower-Level Actions (add new predicates if necessary): Construct new actions that together implement the high-level action.
    - Refer to the suggested decomposition for a LLM-proposed skill sequence. Your actions should cover the skill sequence.
    - Only include actions that result in a state change and are not already defined below # Actions.
    - Do not include meta-actions like confirm, complete, sense, or computation-related steps.
    - Construct new predicates that are required to capture the necessary conditions and effects of the lower-level actions.
    - Define individual PDDL actions for each skill you intend to use.
4. Specify the Goal State: List the predicates that reflect the intended outcome of the high-level action.

{syntax}
\end{lstlisting}
Assistant
\begin{lstlisting}
### Explanation

The high-level action `(grasp-part arm lamp_bulb table)` changes the state from the lamp bulb being on the table to the robot arm holding it. The precondition requires the part to be on the table and not currently held; the effect requires that the robot is holding the part and that the part is no longer on the table.

To decompose this into lower-level PDDL actions, the suggested skill sequence is:

1. hover above the part  
2. set the gripper around the part  
3. close the gripper (to grasp)  
4. lift the part by moving the arm linearly up  

### Change/Add Predicate Definitions

- (hovering_above ?r - robot ?p - part): state. The robot ?r is currently positioned hovering above part ?p.
- (gripper_around ?r - robot ?p - part): state. The robot's gripper is opened and correctly positioned around part ?p.
- (gripper_closed ?r - robot): state. The robot's gripper is closed.
- (holding ?r - robot ?p - part): state. The robot is holding the part with its gripper. (already given)
- (on_table ?p - part ?t - table): state. The part rests on the table. (already given)

### Change/Add Action(s)

1. hover-above-part: add  
    - Description: The robot moves and hovers above the specified part.  
    - PDDL Definition:
        ```pddl
        (:action hover-above-part
            :parameters (?r - robot ?p - part ?t - table)
            :precondition (on_table ?p ?t)
            :effect (hovering_above ?r ?p)
        )
        ```

2. set-gripper-around-part: add  
    - Description: The robot opens its gripper and positions it around the part to grasp. Requires hovering over the part.  
    - PDDL Definition:
        ```pddl
        (:action set-gripper-around-part
            :parameters (?r - robot ?p - part)
            :precondition (hovering_above ?r ?p)
            :effect (and (gripper_around ?r ?p) (not (gripper_closed ?r)))
        )
        ```

3. close-gripper: add  
    - Description: The robot closes the gripper, grasping the part that is around its gripper. Requires the gripper to be around the part.  
    - PDDL Definition:
        ```pddl
        (:action close-gripper
            :parameters (?r - robot ?p - part)
            :precondition (gripper_around ?r ?p)
            :effect (and (holding ?r ?p) (gripper_closed ?r) (not (gripper_around ?r ?p)))
        )
        ```

4. move-linear-up: add  
    - Description: The robot lifts the held part by moving linearly up, causing the part to no longer be on the table. Requires holding the part.  
    - PDDL Definition:
        ```pddl
        (:action move-linear-up
            :parameters (?r - robot ?p - part ?t - table)
            :precondition (holding ?r ?p)
            :effect (not (on_table ?p ?t))
        )
        ```
\end{lstlisting}

\subsection{$\text{LLM}_\text{reasoner}$}

\begin{lstlisting}
You are given a decomposition hierarchy and a record of skills executed in a simulated environment. The last skill has failed during execution. Your goal is to identify why the observed effect of the simulation diverged from the expected effect of that skill. The simulation and skill implementations are correct and fixed - you cannot modify them. Your focus is on reasoning about the planning model and its action decomposition.

---

### Context:
{domain description}

State before (set-gripper-around-part-lowlevel arm lamp_bulb):
(on_table lamp_base table)
(on_table lamp_bulb table)
(on_table lamp_hood table)
(hovering_above arm lamp_bulb)

Decomposition Hierarchy
- (grasp-part arm lamp_bulb table)
    - (hover-above-part arm lamp_bulb table): hover_above_part('lamp_bulb')
    - (set-gripper-around-part arm lamp_bulb)
        - (set-gripper-around-part-lowlevel arm lamp_bulb)

Current Operator
(:action set-gripper-around-part-lowlevel
    :parameters (?r - robot ?p - part)
    :precondition (and (hovering_above ?r ?p) (not (gripper_closed ?r)))
    :effect (gripper_around ?r ?p)
)

Executed Python Skill for PDDL Action `(set-gripper-around-part-lowlevel arm lamp_bulb)`:
set_gripper_around_part('lamp_bulb')

Execution Failure:
Expected Change:
- (gripper_around arm lamp_bulb): False -> True

Ground Truth Change:
- (hovering_above arm lamp_bulb): True -> False
- (gripper_around arm lamp_bulb): False -> True

---

Your Task:
1. Summarize the state before executing the failed action.
1. How did the environment change by executing set_gripper_around_part('lamp_bulb').
2. Identify the cause of the deviation
    - Looking at the decomposition hierarchy, are skills missing before executing set_gripper_around_part('lamp_bulb') to ensure a successful execution (i.e. no collision or undefined behavior)?
        - If so:
            - which skills should have been executed before set_gripper_around_part('lamp_bulb') and can the preconditions be tightened to ensure that?
        - else:
            - should actions be removed?
            - why are the expected effects different and how must the PDDL action set-gripper-around-part-lowlevel be changed to realign it with the ground truth change?: Adapt the PDDL action.
            - should additional predicates be invented to capture the effects more accurately?
            - are preconditions missing in any of the PDDL action definitions?
            - or, were any skills redundant, missing, or incorrectly ordered?
    Note that:
    - We use FastDownward Planning, so naming does not impact the plan, only forming correct PDDL definitions does.
    - Do not assume the current expected effects are correct - critically assess and revise them as needed
    - Adapt the PDDL actions in a general way, e.g. by using conditional effects where applicable (`when`, `forall`, `imply`, ...)
\end{lstlisting}
Assistant response
\begin{lstlisting}
1. **Summary of State Before Executing the Failed Action**

- The robot arm (`arm`) is *hovering above* the `lamp_bulb`.
- The gripper is *not closed*.
- The gripper is *not around* the `lamp_bulb`.
- None of the parts are held.
- All parts (`lamp_base`, `lamp_bulb`, `lamp_hood`) are on the table.
- No parts are aligned, touching, or assembled.
- The robot is *not* hovering above any part except the `lamp_bulb`.
- The gripper is *not around* any part at this point.

2. **How Did the Environment Change by Executing `set_gripper_around_part('lamp_bulb')`**

- The predicate `(gripper_around arm lamp_bulb)` changed from `False` to `True`, as expected.
- Unexpectedly, the predicate `(hovering_above arm lamp_bulb)` changed from `True` to `False`.

This means that while the gripper correctly adjusted to be around the lamp bulb, the robot stopped hovering above it as a side effect.

3. **Identification of the Cause of the Deviation**

- The PDDL action for `set-gripper-around-part-lowlevel` is:

    ```pddl
    (:action set-gripper-around-part-lowlevel
        :parameters (?r - robot ?p - part)
        :precondition (and (hovering_above ?r ?p) (not (gripper_closed ?r)))
        :effect (gripper_around ?r ?p)
    )
    ```

(...)

### **Summary and Recommendation**

- The main reason for the deviation is an incomplete PDDL effect specification in `set-gripper-around-part-lowlevel`.  
- The robot physically must move from a hovering pose to a gripper-encompassing pose, which invalidates `hovering_above` for the part.  
- Add `(not (hovering_above ?r ?p))` to the effects of the action to align the expected effects with actual behavior.  
- No additional skills or predicates are needed, but the planner must be aware that hovering and gripper placement are mutually exclusive states.
\end{lstlisting}
User prompt to get decision of reasoner
\begin{lstlisting}
Determine the most probable fix type based on the following options:
- one of the action definitions listed in `decomposition hierarchy` must be corrected: 'pddl-fix'
- some skills should be executed before the action: 'prior-skills': list the 
- the skill was incorrectly instantiated or used: 'incorrect-instantiation'
- alternatively, the pddl action (set-gripper-around-part-lowlevel arm lamp_bulb) should be implemented with multiple skills: 'multiple-skills'

Independent of the chosen fix type, list ALL operators that must be edited to resolve the issue. If multiple operators must be changed, list them comma-separated ([op1, op2, ...]).

Output Format:
```json
{
    "type_of_fix": "<chosen-fix-type>",
    "operators": ["<corrected-action>", "..."]
}
```
\end{lstlisting}
Final decision of reasoner
\begin{lstlisting}
```json
{
    "type_of_fix": "pddl-fix",
    "operators": ["set-gripper-around-part-lowlevel"]
}
```
\end{lstlisting}

%% file: eval/alg-decomposition.tex
\begin{algorithm}[t]
\caption{Recursive Operator Decomposition in LODGE}
\label{alg:decomposition}
\begin{algorithmic}[1]
\Require Instruction $I$, initial predicates $\Psi_\text{init}$, skill library $\Pi$
\Ensure Hierarchical domain model $\langle \mathcal{D}^{(0)}, \dots, \mathcal{D}^{(d)} \rangle$
\Function{Traverse}{$\mathcal{D}^{(i)}, \mathcal{P}^{(i)}, \Pi$}
    \State Learn classifiers $c_{\psi_\text{new}}$ for new \textit{state-based} predicates $\psi_\text{new}$ in $\mathcal{D}^{(i)}$
    \State Generate plan $a_{1:n}^{(i)} = \text{GenPlan}_\text{LLM}(\mathcal{D}^{(i)}, \mathcal{P}^{(i)})$
    \For{each action $a_j^{(i)}$ in $a_{1:n}^{(i)}$}
        \State Suggest skills $\pi_{0:T-1} = \text{LLM}_\text{translate}(a_j^{(i)}, \Pi, \mathcal{O})$
        \If{length($\pi_{0:T-1}$) = 1}
            \State Motion Validation of $\omega_j^{(i)}$ \Comment{leaf operator}
        \Else
            \State Construct $\mathcal{P}^{(i+1)} = \langle \mathcal{O}, s_j, \text{eff}(s_j, s_{j+1}) \rangle$
            \State Generate $\mathcal{D}^{(i+1)} = \ac{llm}_\text{decomp}(\mathcal{P}^{(i+1)}, \Psi_\text{init}, \Pi, \omega_j^{(i)}, \pi_{0:T-1})$
            \State \Call{Traverse}{$\mathcal{D}^{(i+1)}, \mathcal{P}^{(i+1)}, \Pi$}  \Comment{recursive decomposition}
        \EndIf
    \EndFor
\EndFunction
\State Initialize $\mathcal{D}^{(0)}, g^{(0)} = \text{LLM}_\text{domain}(I, c_\Psi(x_0), \Psi_\text{init}, \Pi)$
\State \Call{Traverse}{$\mathcal{D}^{(0)}, \mathcal{P}^{(0)}, \Pi$}
\end{algorithmic}
\end{algorithm}